\documentclass[letterpaper, 10 pt, conference]{ieeeconf}  

\IEEEoverridecommandlockouts                              

\usepackage{subfiles}
\usepackage{amsmath}
\usepackage{amsfonts}
\usepackage{amsthm,mathrsfs}
\usepackage{bm}
\usepackage{cases}
\usepackage{balance}
\usepackage{mathtools}
\usepackage[urlcolor=blue]{hyperref} 
\usepackage{cleveref}
\usepackage{xcolor}
\usepackage{algorithmic,algorithm}
\usepackage{shortcuts}

\title{\LARGE \bf
Data-Driven Predictive Control for Robust Exoskeleton Locomotion
}

\author{Kejun Li$^{1}$, Jeeseop Kim$^{2}$, Xiaobin Xiong$^{3}$, Kaveh Akbari Hamed$^{4}$, Yisong Yue$^{5}$, Aaron D. Ames$^{2,6}$
\thanks{This research was supported by Wandercraft. Research involving human subjects was conducted under IRB No. 21-0693. The work of Kaveh Akbari Hamed was supported by the NSF under Grants 1923216 and 2024772. The work of Aaron D. Ames was supported by the NSF under Grant 1923239.}
\thanks{$^{1}$K. Li is with the Department of Computation and Neural Systems, Caltech, Pasadena, CA 91125, USA, {\tt\small kli5@caltech.edu}.}\newline
\thanks{$^{2}$J. Kim and A. D. Ames are with the Department of Mechanical and Civil Engineering, Caltech, Pasadena, CA 91125, USA, {\tt\small \{jeeseop, ames\}@caltech.edu}.}\newline
\thanks{$^{3}$X. Xiong is with the Department of Mechanical Engineering, University of Wisconsin-Madison, Madison, WI 53706, USA, {\tt\small xiaobin.xiong@wisc.edu}.}%
\thanks{$^{4}$K. Akbari Hamed is with the Department of Mechanical Engineering, Virginia Tech, VA 24061, USA, {\tt\small kavehakbarihamed@vt.edu}.} 
\thanks{$^{5}$Y. Yue is with the Department of Computing and Mathematical Sciences, Caltech, Pasadena, CA 91125, USA, {\tt\small yyue@caltech.edu}.}%
\thanks{$^{6}$ A. D. Ames is with the Department of Control and Dynamical Systems, Caltech, Pasadena, CA 91125, USA, {\tt\small ames@caltech.edu}.}%
}

\begin{document}

\maketitle
\thispagestyle{empty}
\pagestyle{empty}


\begin{abstract} 
Exoskeleton locomotion must be robust while being adaptive to different users with and without payloads. 
To address these challenges, this work introduces a data-driven predictive control (DDPC) framework to synthesize walking gaits for lower-body exoskeletons, employing Hankel matrices and a state transition matrix for its data-driven model. 
The proposed approach leverages DDPC through a multi-layer architecture. At the top layer, DDPC serves as a planner employing Hankel matrices and a state transition matrix to generate a data-driven model that can learn and adapt to varying users and payloads. At the lower layer, our method incorporates inverse kinematics and passivity-based control to map the planned trajectory from DDPC into the full-order states of the lower-body exoskeleton. We validate the effectiveness of this approach through numerical simulations and hardware experiments conducted on the Atalante lower-body exoskeleton with different payloads. Moreover, we conducted a comparative analysis against the model predictive control (MPC) framework based on the reduced-order linear inverted pendulum (LIP) model. Through this comparison, the paper demonstrates that DDPC enables robust bipedal walking at various velocities while accounting for model uncertainties and unknown perturbations. 
\end{abstract}

\section{INTRODUCTION}
\begin{figure}
  \centering
  \noindent\includegraphics[width=0.97\linewidth]{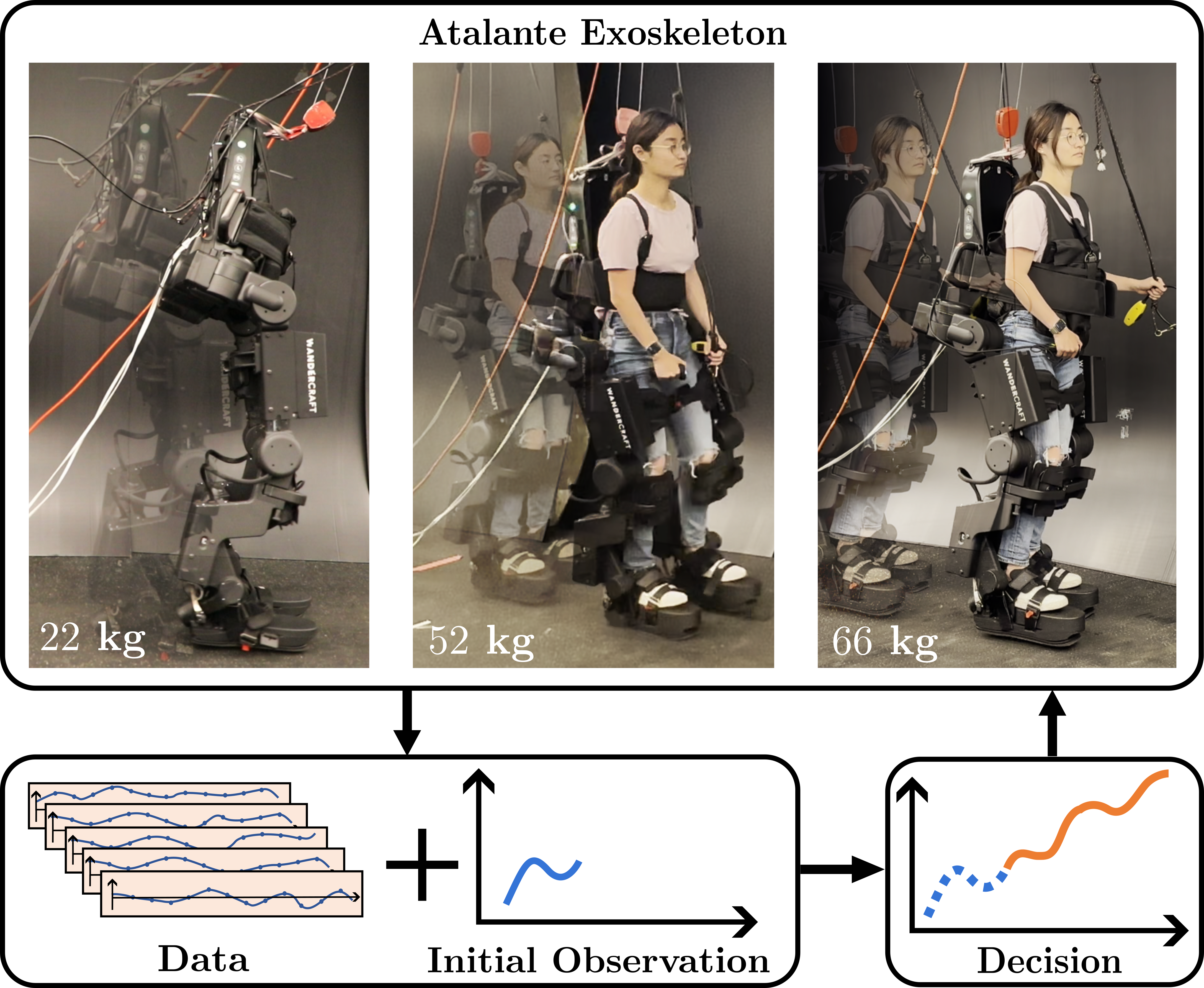}
  \caption{Illustration of data-driven predictive control for bipedal locomotion on lower-body exoskeleton Atalante with various payloads.}
  \label{fig:concept}
  \vspace{-2em}
\end{figure}

Recent research has made substantial advances in lower-body exoskeletons locomotion by enabling stable, dynamic, crutchless walking \cite{agrawal2017first,harib2018feedback}. However, the effort to synergistically integrate users' walking experiences with the functionality of robotic assistive devices is still in progress \cite{tucker2020preference}. The primary challenges hindering this progress include the complex nature of robotic bipedal movement—often described as a high degrees of freedom (DoFs), underactuated hybrid system \cite{grizzle2014models}—along with uncertainties in accurately modeling these systems and the unpredictable dynamics of user-exoskeleton interactions due to unmodeled user movements inside the exoskeleton (see Fig. \ref{fig:concept}).
 




Existing approaches of synthesizing motion for bipedal locomotion usually involve solving trajectory optimization problems using either the robot's full or reduced-order model to find feasible trajectories given the desired walking speed. Methods such as hybrid zero dynamics utilize the full dynamics of the system, offering the advantage of generalizability across platforms and behaviors. It has achieved stable locomotion both on bipedal robots and robotics assistive devices \cite{westervelt2003hybrid,reher2020algorithmic,tucker2021preference, li2022natural}. However, they are computationally demanding, often relying on large-scale offline trajectory optimization, and is prone to convergence issues when seeking stable periodic orbits. For computational benefits, synthesizing motion using a reduced-order model (ROM) such as linear inverted pendulum (LIP) and its variants are appealing since the planning problem depends only on a linear system \cite{kajita2003biped,wensing2013high,englsberger2015three}. Nevertheless, these methods often entail specific model assumptions, such as maintaining a constant center of mass height and zero angular momentum. Choosing the appropriate ROM necessitates careful consideration of the unique characteristics of the robotic platform in use while mitigating any potential discrepancies introduced by these assumptions when designing the desired behaviors. 

With the advancement in modern computational power, pure data-driven approaches, such as reinforcement learning, have offered a model-free approach to train controllers by exploiting large amounts of data from simulators \cite{li2021reinforcement}. While offering robustness, these approaches often require extensive training data and are sensitive to reward design. This motivates leveraging data-driven approaches to learn a more accurate representation of the system dynamics, such as learning the residual dynamics either using Gaussian process \cite{kabzan2019learning} or deep neural network \cite{shi2019neural}, to use in conjunction with classic control methods. In the context of locomotion, due to the high DoFs, to allow for online planning capability, existing work focuses on achieving robust locomotion \cite{dai2023data,xiong2021robust} via learning a reduced-order representation of the system to mitigate model mismatches. 

Another data-driven methodology, representing dynamics system using behavioral systems theory \cite{markovsky2006exact}, has been introduced and proven to be an effective control method for linear time-invariant (LTI) systems. When this methodology is incorporated with predictive control framework, it is referred to as either data-enabled predictive control (DeePC) \cite{coulson2019data} or data-driven predictive control (DDPC). This approach not only has relatively mitigated computation cost and but also demonstrated its efficacy in controlling the underactuated quadrupedal robot \cite{fawcett2022toward} and the multiple quadrupedal robot systems characterized by highly nonlinear interconnected dynamics \cite{fawcett2023distributed,fawcett2023_IEEEAccess}.
Unlike most modern legged robots that operated effectively under the assumption of having negligible legs \cite{xiong20223,hubicki2016atrias}, Atalante has a significant portion of its mass concentrated in the legs (see Fig. \ref{fig:concept}). As a result, we are interested in proposing a data-driven dynamic model and utilizing it for trajectory planning to address these challenges.

\begin{figure}
    \centering
    \includegraphics[width=0.95\linewidth]{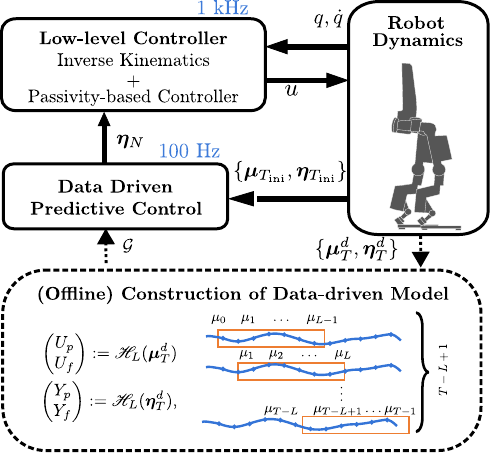}
    \caption{Overview of the proposed layered control framework composed of the DDPC as a planner with constructed data-driven model and low-level controller.}
    \vspace{-2em}
    \label{fig: control_diagram}
\end{figure}
This paper presents a layered framework utilizing data-driven predictive control (DDPC) for trajectory planning and control of the lower-body exoskeleton Atalante as illustrated in Fig. \ref{fig: control_diagram}. The primary contributions of this study are threefold. First, we propose a data-driven dynamic model employing behavioral systems theory with time-domain trajectories of the center of mass (CoM) and center of pressure (CoP), inspired by the LIP model for bipedal locomotion. Second, we design a trajectory planner based on convex DDPC with the proposed data-driven model at the higher level of the proposed layered framework. At the lower level of the framework, the controller incorporates inverse kinematics and passivity-based controllers to translate the planned trajectory from DDPC into the full-order model of the lower-body exoskeleton (Fig. \ref{fig: ROM}a). Third, we conduct experimental evaluations of the proposed data-driven layered framework through numerical simulations and hardware demonstrations. Comparative analysis in simulations demonstrates that the proposed data-driven layered framework effectively stabilizes bipedal gaits at higher speeds, in contrast to the traditional model predictive control (MPC)-based planner for the LIP model. Furthermore, hardware experiments with different kinematics and payloads showcase the ability of this framework to account for user variability as well as the robustness under deviations from nominal data-driven model. 

\section{Preliminaries}

Considering a robotic system with an $n$-dimensional configuration space, we denote the generalized coordinates by $q = \col(q_b,q_a) \in \mathcal{Q} \subset \mathbb{R}^n$, where ``$\col$'' denotes the column operator, $q_b \in SE(3)$ represents the floating-base coordinates and $q_a \in \R^m$ represents the actuated DoFs. The control input is denoted by $u \in \R^m$. The full-order state is denoted by $x = \col (q,\dot{q}) \in T\mathcal{Q}$, where $T\mathcal{Q}$ represents
the tangent bundle of the configuration manifold. 

\subsection{Models of Bipedal Locomotion}
\noindent \textit{\underline{Full-Order Model:}} Bipedal locomotion can be described as a hybrid dynamic system consisting of domains with continuous dynamics and discrete transitions due to impact events \cite{westervelt2003hybrid}. In this work, we only consider a single domain with a single support phase where the stance foot is completely flat on the ground. We assume the ground height is always zero. The single support domain $\mathcal{D}$ and guard $\mathcal{S}$ of the system is defined as $\mathcal{D} = \{x \in T\mathcal{Q}: \zsw(q) \geq 0\}$ and $\mathcal{S} = \{x \in T\mathcal{Q}: \zsw(q) = 0, \dot{\pos}^z_{\textrm{sw}}(q) < 0\}$, where $\textrm{p}_{\textrm{sw}}(q)$ denotes the swing foot position, and the superscript denotes the axis we are considering (i.e., swing foot vertical position). The resulting hybrid system $\mathcal{H}$ can be defined as follows:
\begin{numcases}{\mathcal{H}:}
\dot{x} = f(x) + g(x)\,u & $x \in \mathcal{D} \setminus \mathcal{S}$, \label{eq: continuous_dynamics}
\\
x^+ = \Delta(x^-) & $x^- \in \mathcal{S}$, \label{eq: discretecontrol}
\end{numcases}
where \eqref{eq: continuous_dynamics} denotes the continuous full-order and Lagrangian dynamics, $\Delta: \mathcal{S} \rightarrow \mathcal{D}$ represents the discrete event at foot strike, and the superscripts ``$-$'' and ``$+$'' stand for the instants before and after the impact event, respectively.
\noindent \textit{\underline{Reduced-Order Inverted Pendulum Model:}} 
Since we only consider flat-footed walking in this work, we can assume all contact points, denoted by $p_i$ for the contacting index $i$, are on the flat ground, that is, $p^z_i = 0$. By constraining angular momentum to be constant (i.e., $\dot{\bm{L}}_{\textrm{com}} = 0$), one can simplify the Newton-Euler equation for Centroidal dynamics \cite{wieber2016modeling} as follows: 
\begin{align}
    \comacc^{\{x,y\}} = \frac{\comacc^z+g}{\compos^z}
    \left(\compos^{\{x,y\}} - \coppos^{\{x,y\}}\right),
    \label{eq:variableip}
\end{align}
where $g$ is the gravitational constant, $\coppos^{\{x,y\}} := \frac{\sum_i f^z_i p_i^{\{x,y\}}}{\sum_i f_i^z}$ represents the CoP position, $\compos$ denotes the CoM position, and $\comacc$ represents the CoM acceleration. Since the contact forces $f_i$ are unilateral with $f_i^z \geq 0$, the CoP has to lie inside the convex hull of the contact points, that is, $\coppos \in \text{conv}\{p_i\}$. If we additionally assume constant vertical CoM position $\compos^z$, one could simplify \eqref{eq:variableip} to obtain the classic LIP model that is widely used for motion synthesis in bipedal locomotion \cite{kajita2002realtime}. Motivated by the relation between the CoP inputs and CoM outputs in the LIP model, we will develop a data-driven model in the next sections.  

\subsection{Behavioral Systems Theory}

This section briefly reviews some of the fundamental results of the behavioral systems theory. Let us consider a state representation of a discrete-time LTI system as follows:
\begin{equation}
 \begin{aligned}
   & \theta(t+1) &&= A\,\theta(t) + B\,\mu(t) \\
   & \eta(t)     &&= C\,\theta(t) + D\,\mu(t),
 \end{aligned} \label{eq: LTI}
 \end{equation}
where \( \theta(t) \in \mathbb{R}^\beta \), \( \mu(t) \in \mathbb{R}^\kappa \), \( \eta(t) \in \mathbb{R}^\nu \) represent the state vector, control inputs, and outputs, respectively, at time \( t \in \mathbb{Z}_{\geq 0} := \{0,1,\dots \} \), and \( A \in \mathbb{R}^{\beta \times \beta} \), \( B \in \mathbb{R}^{\beta \times \kappa} \), \( C \in \mathbb{R}^{\nu \times \beta} \), \( D \in \mathbb{R}^{\nu \times \kappa} \) denote the unknown state matrices. In behavioral systems theory, a dynamical system is defined as a 3-tuple $(\mathbb{Z}_{\geq 0},\mathbb{W},\mathscr{B})$, where $\mathbb{W}$ is a signal space and $\mathscr{B} \in \mathbb{W}^{\mathbb{Z}_{\geq 0}}$ is the behavior. In contrast with classical systems theory with a particular parametric system representation such as that of \eqref{eq: LTI}, behavioral systems theory focuses on the subspace of the signal space where system trajectories live.  Let $\bm{\mu}_T := \col(\mu_0, \mu_1,\dots,\mu_{T-1})$ be an input trajectory with length $T \in \mathbb{N}:=\{1,2,\cdots\}$ applied in $\mathscr{B}$ with the corresponding output trajectory $\bm{\eta}_T := \col(\eta_0, \eta_1,\dots,\eta_{T-1})$. We can construct the Hankel matrix with $\bm{\mu}_T$ by concatenating trajectory with length $L \in \mathbb{N}$ and $T > L$ as follows:
\[
\Hankel{L}{\bm{\mu}_T} := 
\begin{bmatrix}
\mu_0 & \cdots & \mu_{T-L} \\
\vdots & \ddots & \vdots \\
\mu_{L-1} & \cdots & \mu_{T-1}
\end{bmatrix} \in \mathbb{R}^{\kappa L \times (T-L+1)}.
\]  
Here we note that the Hankel matrix with $\bm{\eta}_T$, $\Hankel{L}{\bm{\eta}_T}$, can be constructed analogously.
The signal $\bm{\mu}_T$ is said to be persistently exciting of order L if $\Hankel{L}{\bm{\mu}_T}$ is of full row rank \cite{Berberich_DDPC}. If a given data input-output (I-O) trajectory of the system, denoted by the pair $(\bm{\mu}^{d}_{T},\bm{\eta}^{d}_{T})$, is persistently exciting of order $L+\beta$, from Fundamental Lemma \cite{Berberich_DDPC, willems2005note}, any trajectory of the LTI system can be constructed using a linear combination of the columns of the Hankel matrices. In particular, one can use the columns of the Hankel matrices to develop a data-driven model to predict the system's future behavior. 

To make this notion more precise, let us take $L$ as $L = T_{\textrm{ini}} + N$, where $T_{\textrm{ini}}$ and $N$ denote the estimation horizon and control horizon, respectively. The estimation horizon is used to estimate the system's initial state from past I-O measurements. The control horizon is used for the predictive controller. We can partition the Hankel matrices into the past and future portions accordingly as follows: 
\begin{align*}
\begin{bmatrix}
U_p \\
U_f
\end{bmatrix}
:= \Hankel{L}{\bm{\mu}_T^d} \quad \quad
\begin{bmatrix}
Y_p \\
Y_f
\end{bmatrix}
:= \Hankel{L}{\bm{\eta}_T^d},
\end{align*}
where $U_p \in \R^{\kappa T_{\textrm{ini}} \times (T-L+1)}$, $U_f \in \R^{\kappa N \times (T-L+1)}$, $Y_p \in \R^{\nu T_{\textrm{ini}} \times (T-L+1)}$,$Y_f \in \R^{\nu N \times (T-L+1)}$. From the Fundamental Lemma \cite{coulson2019data, Berberich_DDPC, willems2005note}, any new trajectory lies in the range space of the Hankel matrices, or equivalently, there exists $\gamma\in\mathbb{R}^{T-L+1}$ such that 
\begin{align}
\begin{bmatrix}
U_p \\
Y_p \\
U_f \\
Y_f
\end{bmatrix} \gamma = \begin{bmatrix}
\bm{\mu}_{T_{\textrm{ini}}} \\
\bm{\eta}_{T_{\textrm{ini}}} \\
\bm{\mu}_N \\
\bm{\eta}_N
\end{bmatrix},
\label{eq:traj_constraint}
\end{align}
where $\bm{\mu}_{T_{\textrm{ini}}}$ and $\bm{\eta}_{T_{\textrm{ini}}}$ denote the past portions of the I-O trajectories over the estimation horizon of $T_{\textrm{ini}}$. Similarly, $\bm{\mu}_{N}$ and $\bm{\eta}_{N}$ represent the predicted (i.e., future) portions of the I-O trajectories over the control horizon of $N$. Since the dimensionality of the $\gamma$ vector in formulating predictive controllers is huge, we aim to remove $\gamma$ from \eqref{eq:traj_constraint}. One way is to have an offline approximation for $\gamma$ using least squares similar to \cite{fawcett2022toward, PowerSysDeePC} to obtain the data-driven model as
\begin{equation}
    \bm{\eta}_{N}= Y_{f}\gamma = \underbrace{Y_{f}\begin{bmatrix}
        U_{p}\\
        Y_{p}\\
        U_{f}
    \end{bmatrix}^{\dagger}}_\mathcal{G}
    \begin{bmatrix}
        \bm{\mu}_{T_{\textrm{ini}}}\\
        \bm{\eta}_{T_{\textrm{ini}}}\\
        \bm{\mu}_{N}
    \end{bmatrix},
\end{equation}
where $(\cdot)^\dagger$ represents the pseudo inverse and $\mathcal{G}$ denotes the data-driven state transition matrix over $N$-steps.

\begin{figure}
    \centering
\includegraphics[width=0.95\linewidth]{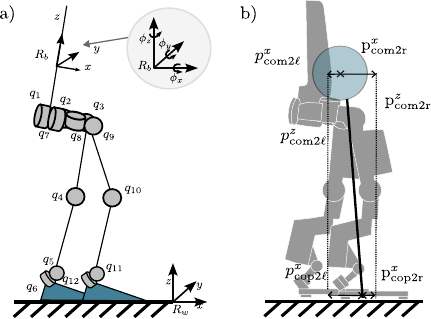}
    \caption{a) Generalized coordinates for the lower-body exoskeleton Atalante. b) The input and output variables in the x-direction to be used for the Hankel matrix construction.}
    \vspace{-2em}
    \label{fig: ROM}
\end{figure}

\section{Data-Driven Motion Planner}

This section aims to present the proposed DDPC-based trajectory planner at the high level of the control scheme for bipedal locomotion (see Fig. \ref{fig: control_diagram}). 

\subsection{Construction of the Data-Driven Model} Assuming reasonable behavior for the actuated coordinates, the difficulty and complexity of the bipedal locomotion usually lie in the control and planning for the weakly actuated or underactuated Centroidal states. In this context, designing the CoM trajectory encapsulated all the requisite DoFs' information, although their individual dynamics are not described explicitly. Hence instead of using the full-order states $x$ for trajectory planning, we are focusing on the Centroidal states to construct the Hankel matrices. Specifically, we draw inspiration from the LIP model that considers the CoM and CoP of the robot. We consider a local representation of these states with respect to either the left or right foot frame as $\coptoi^{x,y} := \coppos^{x,y} - \textrm{p}^{x,y}_{\ell/\textrm{r}}$ and $\comtoi^{x,y,z} := \compos^{x,y,z} -\textrm{p}^{x,y,z}_{\ell/\textrm{r}}$ as shown in Fig. \ref{fig: ROM}b.



An intuitive choice would be using the CoM and CoP trajectories in the stance-foot frame. However, this choice would create discontinuity during domain switches, posing challenges for planning through impact. To prevent the state space from monotonically increasing during forward walking and to maintain continuity in the trajectory planning, we adopt a redundant representation where the trajectory of the CoM and CoP with respect to both stance and swing feet are being considered. Hence, the input $\mu \in \R^4$ and output $\eta \in \R^6$ of the date-driven model are chosen as follows:
\begin{align*}
    \eta := \col(\comtoL^{x,y,z}, \, \comtoR^{x,y,z}) \quad \quad \mu := \col(\coptoL^{x,y}, \, \coptoR^{x,y}).
\end{align*}
Notably, with this redundant formulation, we also provide trajectories in the following domain throughout the prediction horizon for the low-level controller to track in case of unexpected early or late impact. This is especially advantageous in scenarios involving uncertain impact timings or unexpected communication delays, significantly enhancing the framework's practical applicability. We do not include velocity terms in our representation as the velocity and acceleration information is implicitly captured in the position trajectory in the Hankel matrix. 
We remark that the proposed data-driven model, inspired by the LIP model, captures more comprehensive information about the system with implicit consideration of swing foot trajectory and the effect of the low-level whole-body controller on the system dynamics. A comparative analysis of the performance between these two models and their corresponding planner architectures will be presented in Section \ref{sec: result}.



\begin{figure*}
    \centering
    \includegraphics[width=\linewidth]{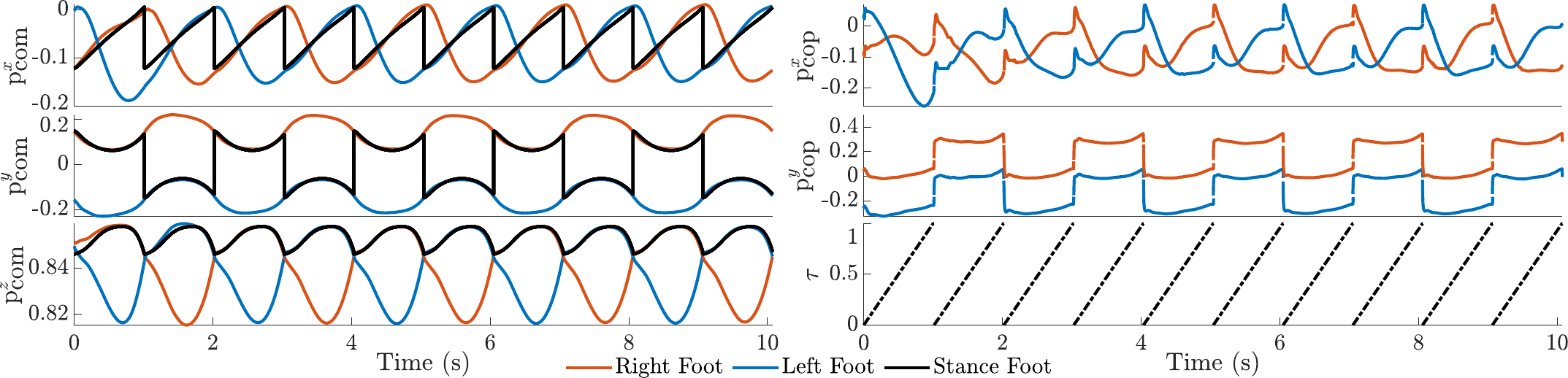}
    \vspace{-1.5em}
    \caption{One set of the planned CoM and CoP trajectories from the DDPC planner and the tracked trajectory in simulation in right foot frame, the left foot frame trajectories, and the Stance foot. The stance foot frame trajectory in black is generated from DDPC. The corresponding phase variables are plotted in the dashed line.}
    \label{fig: sim_traj}
    \vspace{-1em}
\end{figure*}

\begin{figure*}
    \centering
    \includegraphics[width=\linewidth]{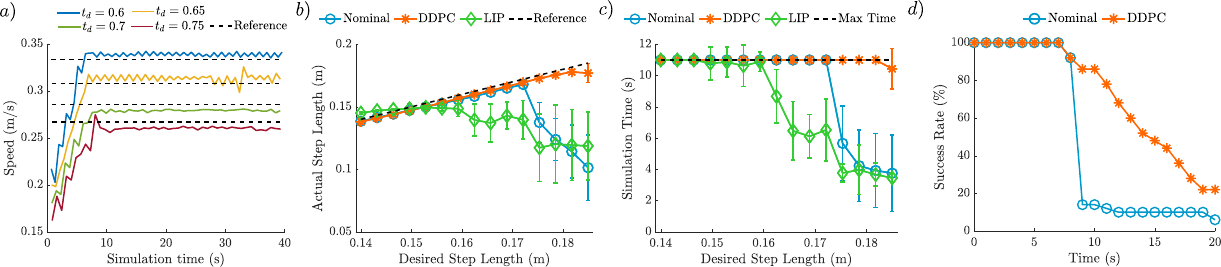}
    \vspace{-1.5em}
    \caption{Simulation comparison over nominal indicated by blue circles, DDPC indicated by orange stars, and MPC indicated by green diamonds. a) DDPC planner planning trajectories for increasing desired speed, capped at maximum step length $0.2$ m at different step duration $t_d$. b) tracking performance over different desired step length with the same step duration. The dashed line indicate the ideal performance. Error bar indicates the standard deviation over 50 models. c) Simulation time before robot falling for the tracking performance comparison. Error bar indicates the standard deviation over 50 models.  The maximum simulation time is $11$ s, indicated by the horizontal dash line. d) Comparison of Nominal and DDPC under time-varying perturbation applied on the negative $x$ direction.}
    \label{fig:sim_result}
    \vspace{-1.5em}
\end{figure*}

\subsection{DDPC Algorithm for Trajectory Optimization} We are now positioned to present the DDPC-based trajectory planner for optimizing the I-O (i.e., CoP and CoM) trajectories. The real-time DDPC planner is formulated as the following strictly convex quadratic program (QP)
\begin{align}\label{eq:deepc_planner}
& \underset{(\bm{\mu}_{N},\bm{\eta}_{N})} \min
& & \sum_{k=0}^{N-1} \left(\|\eta_k - r^\eta_{k}\|_Q^2 + \|\mu_k - r^\mu_{k}\|_R^2\right) \notag\\
& \text{subject to}
& & \bm{\eta}_N = \mathcal{G} \begin{bmatrix}
\bm{\mu}_{T_{\textrm{ini}}} \\
\bm{\eta}_{T_{\textrm{ini}}} \\
\bm{\mu}_N
\end{bmatrix} \quad (\textrm{Data-Driven Model})\\
&&& \mu_k \in \mathcal{U}, \quad \eta_k \in \mathcal{P}, \quad k = 0, \ldots, N-1, \notag
\end{align}
where \( \bm{r}^\eta := \col (r^\eta_0, \cdots, r^\eta_{N-1}) \) represents a reference output (i.e., CoM) trajectory, \( \bm{r}^\mu := \col (r^\mu_0, \cdots, r^\mu_{N-1}) \) denotes a reference input (i.e., CoP) trajectory, and \( (\bm{\mu}_{T_{\textrm{ini}}}, \bm{\eta}_{T_{\textrm{ini}}})\) represents the pair of past actual I-O trajectories (i.e., feedback to the DDPC). In addition,  \( \mathcal{U} \subseteq \mathbb{R}^4 \) and \( \mathcal{P} \subseteq \mathbb{R}^6 \) denotes the input and output feasibility sets, respectively. Finally, \( Q \in \mathbb{R}^{6 \times 6} \) and \( R \in \mathbb{R}^{4 \times 4} \) are chosen as positive definite weighting matrices.

\section{Layered Control Framework for Atalante}

In this section, we introduce the remaining components and details for practically implementing the entire layered control framework used to realize locomotion on the lower-body exoskeleton Atalante (see Fig. \ref{fig:concept}). Atalante contains 12 actuated DoFs as illustrated in Fig. \ref{fig: ROM}a, with three actuators on the hip, one on the knee, and two on the ankle for each leg. The device weighs $82$ kg and has adjustable thigh and shin length to account for the varying heights of users. Depending on the physical parameters of the users, the largest kinematically feasible step length in $x$ direction that the device is capable to achieve for flat-footed walking is less than $0.2$ m.

\subsection{Trajectory Planner with DDPC}
We generate a reference trajectory with the open-source toolbox FROST \cite{hereid2017frost} using the full-order system to account for kinematics and dynamics feasibility. This trajectory is described by $7$-th order B\'ezier polynomials with the coefficient matrix of $\alpha = [\alpha_{\textrm{com}}, \alpha_{\phi},\alpha_{\textrm{sw}}]$, where $\alpha_{\textrm{com}}$ describes the CoM trajectory, $\alpha_{\phi}$ described the pelvis orientation and $\alpha_{\textrm{sw}}$ described the swing foot position and orientation. The B\'ezier polynomials are evaluated based on a time-based phase variable. Specifically, considering a desired step duration $t_{d}$ and the initial time at the beginning of the domain $t_0$, we can calculate the phase variable $\tau_k = \frac{t_k-t_0}{t_{\textrm{d}}}$ at time point $t_k$. The reference for the CoM trajectory is then determined by $r^\eta_k = \col(\compos^{x,y,z}(\alpha,\tau_k)-p_{\ell}^{x,y,z}(\alpha,\tau_k), \,\, \compos^{x,y,z}(\alpha,\tau_k)-\pos_{\textrm{r}}^{x,y,z}(\alpha,\tau_k))$. In addition, the reference CoP trajectory is generated via $r^\mu_k = \col(-\pos_{\textrm{sw}}^{x,y}(\alpha,\tau_k),0,0)$ for the right stance and $r^u_k = \col(0,0,-\pos_{\textrm{sw}}^{x,y}(\alpha,\tau_k))$ for the left stance.

Without explicit guidelines to construct the Hankel matrix for nonlinear systems, we empirically determine the hyperparameters of the DDPC algorithm via a grid search over the space of the discrete-time interval between nearby points $\delta_t \in [0.01,0.03]$, the trajectory length of $T \in [50,600]$, the initial trajectory length of $T_{\textrm{ini}} \in [5,50]$, and the control horizon of $N \in [10,300]$. This search space is constructed considering the computation speed, low-level controller frequency, noisy level of the data, and the amount of data required. We choose $T=400$, $T_{\textrm{ini}} = 10$, $N = 20$, and $\delta_t = 0.02$ with a selection criteria on the accuracy of the least-square approximation over some unseen trajectory. In total, $8$ s of data are used for the Hankel matrix construction, and a trajectory for $0.4$ s is planned. This means our DDPC-based trajectory planner primarily acts as a short-term regulator to stabilize the system. We remark that even though $\pos_{\textrm{com}}^z$ is planned, it is not being used by the low-level controller but only used as part of the states to determine system dynamics. The trajectories are planned at $100$ Hz, which is faster than the interval $\delta_t$ specified. However, since the reference generation is based on the current domain and phase variable, the trajectory planner still has the ability to regulate the system behavior during this interval till the next time sample at which $\bm{\mu}_{T_{\textrm{ini}}}$ and $\bm{\eta}_{T_{\textrm{ini}}}$ are updated.

\subsection{Output Synthesis}
The desired walking behavior is encoded by the task space output $\mathbf{y} = \mathbf{y}^{\text{act}} - \mathbf{y}^{\text{des}}$, where $\mathbf{y}^{\text{act}} \in \R^{12}$ and $\mathbf{y}^{\text{des}} \in \R^{12}$ are the actual and desired outputs, respectively. In particular, we choose the following outputs for the system:
\begin{align*}
    \mathbf{y}&^{\text{act}} =  \begin{bmatrix}
    \composst^{x,y,z}(q) &  \phi^{x,y,z}_{\textrm{pelv}}(q) & \pos_{\textrm{sw}}^{x,y,z}(q) & \phi^{x,y,z}_{\textrm{sw}}(q)
    \end{bmatrix}\\
    \mathbf{y}&^{\text{des}} = \begin{bmatrix}
    \composst^{x,y,z} &  \phi^{x,y,z}_{\textrm{pelv}}(\alpha) & \pos_{\textrm{sw}}^{x,y,z}(\alpha,\lambda^{x,y}) & \phi^{x,y,z}_{\textrm{sw}}(\alpha)
    \end{bmatrix},
\end{align*}
where the desired COM position $\composst^{x,y}$ is generated by the high-level DDPC planner, and the other desired components are taken as B\'ezier polynomials with the coefficient matrix of $\alpha$ and the step length of $\lambda^{x,y}$. More specifically, the coefficients of pelvis orientation $\phi_{\textrm{pelv}}$ and swing foot orientation $\phi_{\textrm{sw}}$, and $z$-height of CoM and swing foot trajectory are fixed and from the aforementioned reference trajectory. The swing foot $x,y$ trajectories are determined by B\'ezier polynomials connecting the swing foot position at the beginning of the domain (i.e., post-impact state) and the desire foot targets, i.e., $\pos_{\textrm{sw}}^{x,y}(\tau) = (1-\beta(\tau))\,\pos_{\textrm{sw}}(q^+) + \beta(\tau)\,\lambda^{x,y}$, where $\beta$ is a phase-based weighting function. 


\subsection{Low-Level Feedback Controller} Our low-level controller, implemented in C++, receives trajectories for the CoM position target from the DDPC-based trajectory planner. Given that the controller operates at $1$ kHz, much faster than the planner's replan frequency and the discrete-time interval, CoM target positions and velocities for each control tick is obtained via linear interpolation. Subsequently, we employ a Newton-Raphson numerical inverse kinematics (IK) algorithm to calculate the desired joint targets, which are then tracked using a passivity-based control method \cite{gong2022zero}. The kinematics and dynamics evaluation is performed with Pinocchio \cite{carpentier2019pinocchio}. To account for uncertainty in state estimation and impact time, we switch to the next stance domain only when the swing foot ground reaction force exceeds some pre-defined threshold instead of time-based switching.


\begin{figure*}
    \centering
    \includegraphics[width=\linewidth]{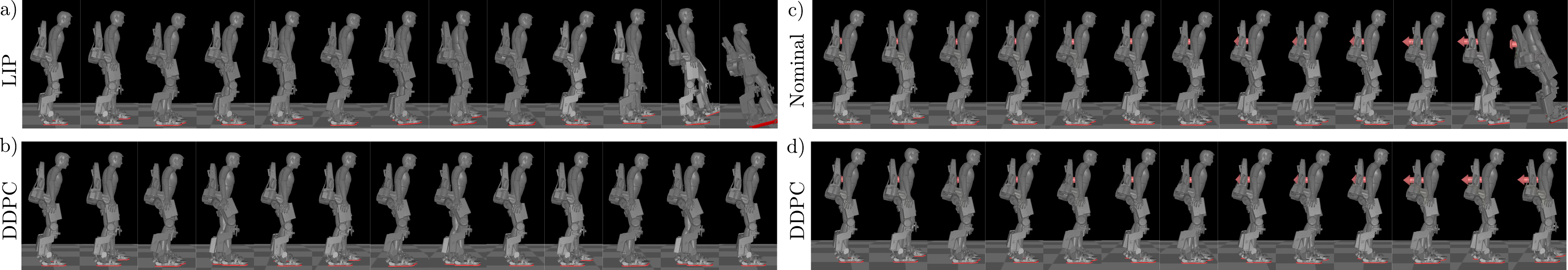}
    \caption{Gait tiles for simulation. Simulations with a) LIP-based MPC and b) DDPC controllers for walking at the speed of $0.16$ (m/s), c) Nominal trajectory and d) DDPC under time-varying perturbations.}
    \label{fig: sim_gait_tile}
    \vspace{-1.5em}
\end{figure*}

\begin{figure}
    \centering
    \includegraphics[width=\linewidth]{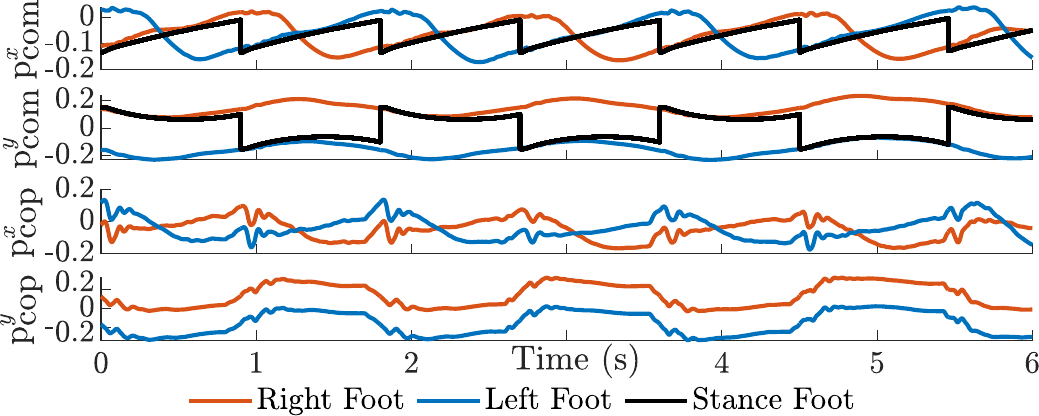}
    \caption{Desired trajectories from the DDPC-based trajectory planner (black solid line) and actual evolving CoM/CoP states from the hardware experiment.}
    \label{fig:hardware_traj}
    \vspace{-2em}
\end{figure}

\section{Experimental Validations} \label{sec: result}
In this section, we present the numerical simulation and hardware experiment results for our proposed framework. The experiment video could be found in \cite{video}. 

\subsection{Simulation Results}
We validate the effectiveness of our framework with numerical simulations in MuJoCo \cite{todorov2012mujoco}. To account for uncertainty in mass distribution estimation for human users, we generated 50 randomized models with identical user total mass by varying the CoM offset and inertia properties, resulting in a $\compos$ ranging from $[-0.122,-0.106]$ m in nominal standing configuration. The low level controller is using the information from the same nominal model constructed with parameters from \cite{winter2009biomechanics}.

\noindent \textit{\underline{Tracking Performance}}: We compare tracking performance over these randomly generated models. For each model, simulation data with different desired step lengths ranging from $0.1$ to $0.15$ m were collected to construct the $\mathcal{G}$ matrix. An example trajectory generated by the DDPC planner and the corresponding actual CoM and CoP states are shown in Fig. \ref{fig: sim_traj}. Moreover, DDPC is able to achieve stable walking in various speed as described in Fig. \ref{fig:sim_result}a. Additionally, we implemented an MPC planner based on the LIP dynamics for comparative analysis. This MPC shares a similar problem structure to that described in \eqref{eq:deepc_planner}, but it substitutes the Hankel matrix trajectory constraints with those of the LIP dynamics. Considering the kinematic limits and feasibility concerns, we opted for a $0.8$ m CoM z-position in the LIP model (as opposed to $0.85$ m) and set $\delta_t = 0.01$ s to discretize the LIP dynamics. The time horizon for the MPC is also chosen as $N_{\textrm{LIP}} = 300$.

The DDPC planner's tracking performance was evaluated against that of the MPC and a nominal trajectory for desired walking speeds between $0.14$ and $0.19$ m/s, with a constant step duration of $1$ s. As shown in Fig. \ref{fig:sim_result}b, the DDPC planner reliably achieves close tracking of the desired step length. As the desired speed increases by increasing the desired step length, the validity of LIP dynamics starts to degrade. In particular, the LIP MPC's performance begins to decline at speeds above $0.16$ m/s, and the nominal trajectory cannot be appropriately tracked beyond $0.17$ m/s, as described in Fig. \ref{fig:sim_result}b. Furthermore, the DDPC planner maintains the system's stability for a longer duration than the other controllers as demonstrated in Fig. \ref{fig:sim_result}c. Here, we define a robot maintaining CoM above a certain threshold throughout the maximum simulation time as a success. It is a failure if QP or IK becomes infeasible or CoM falls below a specified threshold. An example of failure case is shown in Fig. \ref{fig: sim_gait_tile}a for MPC.
Since the DDPC-generated trajectory is based on the feasible trajectory used to construct $\mathcal{G}$, it tends to run into kinematics issues less frequently than physics-based reduced-order model like LIP, where no full-order model system information is exposed to the planner.

\begin{figure*}
  \centering
  \noindent\includegraphics[width=\linewidth]{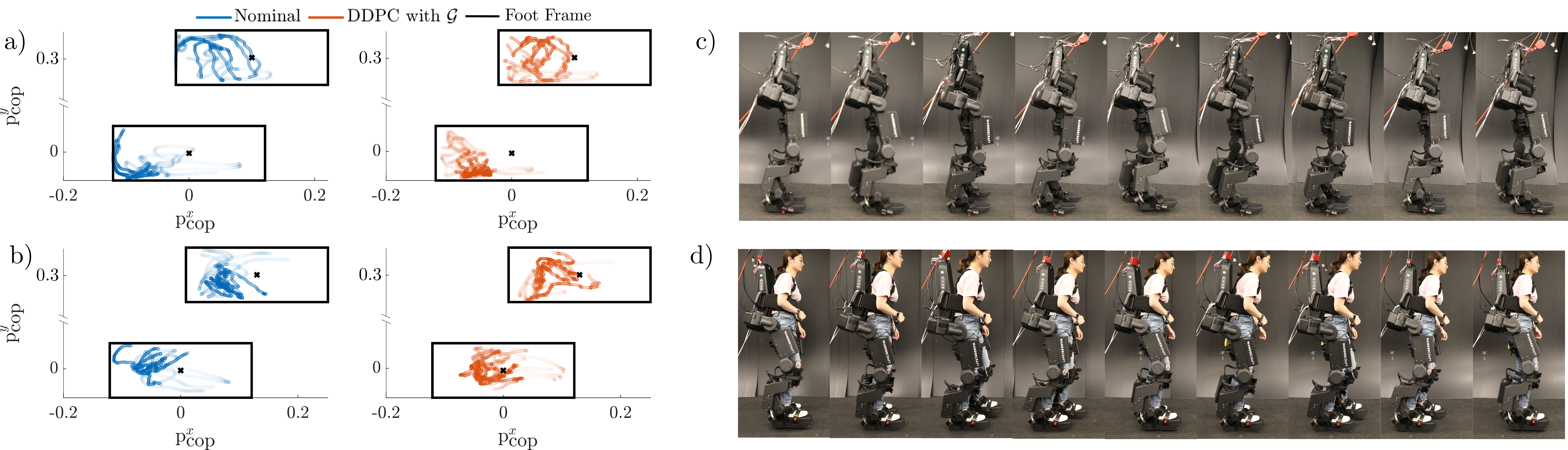}
  \caption{The system evolution with the data-driven layered framework is shown in orange, and the system evolution with nominal trajectory is shown in blue.  a) Experiment result for exoskeleton carrying $20$ kg of payload. CoP position in the foot frame for 5 left stance foot steps and 5 right stance foot steps. b) CoP position for exoskeleton with user inside. c) Gait tiles with the DDPC planner for the $20$ kg payload experiment. d) Gait tiles with the DDPC planner for the experiment with user.}
  \label{fig:Nominal_hardware}
  \vspace{-1em}
\end{figure*}

\begin{figure*}
  \centering
  \noindent\includegraphics[width=\linewidth]{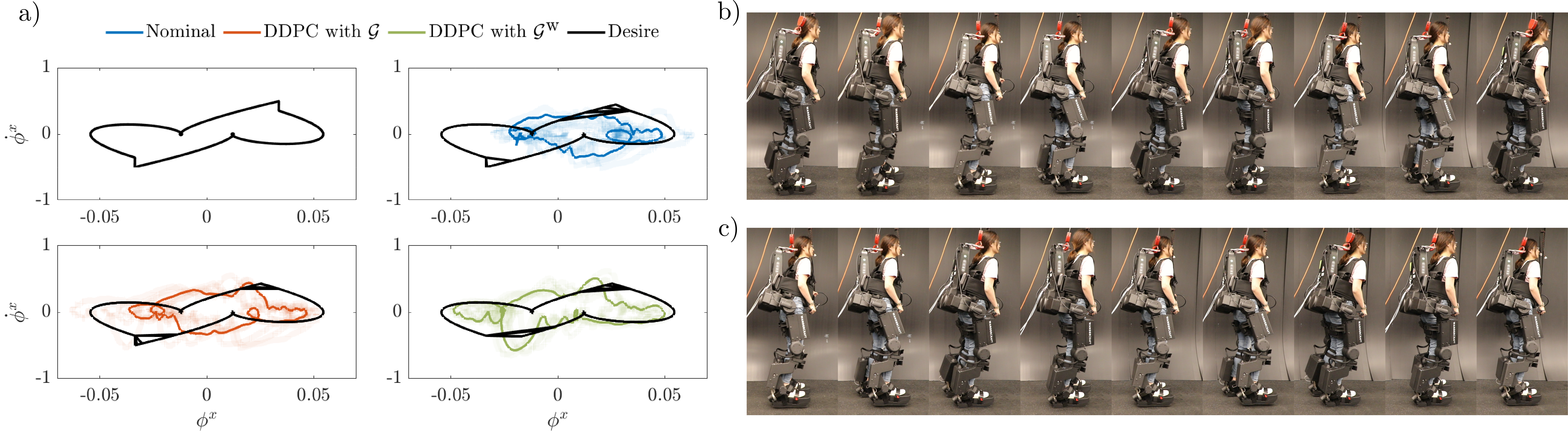}
  \vspace{-1.5em}
  \caption{a) Phase portrait over pelvis roll $\phi_x$ and $\dot{\phi}_x$ for hardware experiment with user wearing additional weight for nominal, DDPC with $\mathcal{G}$, DDPC with $\mathcal{G}^\text{w}$ b) Gait tiles for experiment with DDPC controller with $\mathcal{G}$ with additional weight c) Gait tiles for experiment with DDPC controller with $\mathcal{G}^{\text{w}}$ with additional weight}
  \label{fig:Gait_Tile_weight}
  \vspace{-2em}
\end{figure*}

\noindent \textit{\underline{Adaptation to Perturbation}}: We further evaluate the DDPC's robustness against time-varying external disturbances (see Fig. \ref{fig: sim_gait_tile}c and \ref{fig: sim_gait_tile}d). Specifically, we continuously apply an external perturbation force to the torso's negative $x$ direction. This direction is chosen because the device is more sensitive towards perturbation applied along this axis with its fairly negative nominal CoM position.  The initial perturbation force was set at $5$ N with a discrete increase of $3$ N every $3$ s.  To adapt to this time-varying perturbation, the Hankel matrix is updated online. The choice of the Hankel matrix size requires a balance between satisfying persistently exciting requirements and using obsolete data. The simulation's first $5$ s were used for data collection, which was a smaller trajectory length $T=250$, and $\mathcal{G}$ is reconstructed every $1.5$ s. The least-square approximation step is completed within $70-100$ ms. We ran the same setup for the aforementioned 50 randomly generated models and compared it with the nominal trajectory at $0.13$ m/s desired speed. The DDPC has a higher success rate, defined by the percentage of the models being upright, compared to the nominal trajectory (see Fig. \ref{fig:sim_result}d). 

\subsection{Hardware Results}
The same C++ low-level controller used in simulation, with additional code to interface with Wandercraft API, is run directly on the Atalante onboard computer (i5-4300U CPU @1.90GHz with 8GB RAM at 1kHz). The DDPC planner is run on an external PC (i7-8700K CPU @3.70GHz) communicating with the onboard computer via a UDP network. To account for package delay, a segment of the desired trajectory is sent, and the low-level controller finds the closest discrete target point given the current phase variable. As a result, the low-level controller receives planned trajectory with the upcoming time stamps and could handle the delay caused by planning computation time and UDP communication. An example planned trajectory of CoM and CoP under this setup is shown in Fig. \ref{fig:hardware_traj}. 

\noindent\textit{\underline{Different User Settings}}: We conducted two sets of experiments to test the framework with different payloads and kinematics as this platform is designed to be used by different users. The data collection part for the hardware experiment is similar to that of the simulation setup for tracking performance comparison with gaits at different step lengths. The first set of experiments is conducted with the exoskeleton carrying $20$ kg of payload with the link lengths of a subject of $1.74$ m. DDPC planner is able to regulate the CoP position to a more centered position as depicted in Fig. \ref{fig:Nominal_hardware}a and \ref{fig:Nominal_hardware}b. A different set of hyperparameters with $T=800$, $T_{\textrm{ini}} = 20$, $N = 100$, and $\delta_t = 0.015$ is used. 
The second experiment is conducted with a user of height $1.63$ m and $52$ kg with the same hyperparameter to construct the Hankel Matrix as in tracking performance case. The gait tiles for the experiment with different payloads are shown in Fig. \ref{fig:Nominal_hardware}c and \ref{fig:Nominal_hardware}d.

\noindent\textit{\underline{Uncertain User Mass}}: To investigate the effect of the planner's capability to handle the uncertain mass of the user or in case of the user wanting to carry additional payload, we also conducted a set of experiments with the user carrying an additional $14$ kg of weight (see Fig. \ref{fig:Gait_Tile_weight}). We test both DDPCs with $\mathcal{G}$ constructed from data without carrying the additional weight and with $\mathcal{G}^{\text{w}}$ which is constructed with trials where the user is carrying additional weight. From the phase portrait described in Fig. \ref{fig:Gait_Tile_weight}a, we could see that the DDPC with $\mathcal{G}^{\text{w}}$ resembles a desired limit cycle compared to the DDPC with $\mathcal{G}$ and the nominal controller. The gait tiles for the experiments with $\mathcal{G}$ and $\mathcal{G}^{\text{w}}$ are shown in Fig. \ref{fig:Gait_Tile_weight}b and \ref{fig:Gait_Tile_weight}c, respectively. We also evaluate the cumulative tracking error over $t_{\textrm{total}} =10$ s for output other than the CoM position and yaw related, denoted by $\mathbf{y}_{\text{part}}$, as the CoM positions are different across controllers and the yaw related output is not well approximated due to IMU drifting via evaluating $e = \int_{0}^{t_{\textrm{total}}} \| \mathbf{y}^{\text{act}}_{\text{part}}(t) - \mathbf{y}^{\text{des}}_{\text{part}}(t) \|^2 dt$. This value for the DDPC with $\mathcal{G}$, $\mathcal{G}^{\text{w}}$, and nominal is $0.4627$, $0.4319$, and $0.5181$, respectively.

\section{CONCLUSIONS}

This paper successfully demonstrated the DDPC framework's application on lower-body exoskeleton, both in simulations and on hardware. Through detailed simulation analyses, the DDPC framework proved its effectiveness in stabilizing the system, surpassing traditional physics-based template models such as LIP, particularly at increased desired speeds. The framework's robustness was also validated on hardware, showcasing its ability to accommodate model discrepancies beyond the initial model used for data collection. Furthermore, we introduced a time-varying perturbation in the simulation while updating the transition matrix online. Although further research is necessary to refine the online update process of the Hankel matrices systematically, these results underscore the framework's capacity to adapt to changing environments. However, it is crucial to note that the DDPC planner's performance is intrinsically linked to the trajectory used to construct the Hankel matrices. It can only enhance and build upon the capabilities of the nominal controller used for data collection. Moreover, considering the inherent limitations in the actuation of the CoM horizontal position, future studies will explore how to extend the proposed framework to incorporate foot placement and step timing planning to enhance stability and performance. Additionally, further investigation into incorporating more information on user movement would be beneficial.


\small{\section*{ACKNOWLEDGMENTS}
The authors would like to thank the entire Wandercraft team for their continued guidance and technical support with Atalante.}
\bibliographystyle{IEEEtran}
\balance
\bibliography{IEEEabrv, References}


\end{document}